# Unsupervised Sparse Coding-based Spiking Neural Network for Real-time Spike Sorting

**Alexis Mélot [1,2], Sean U.N. Wood[2], Yannick Coffinier[1], Pierre Yger[3] and Fabien Alibart[1,4]**
[1] Institut d'Électronique, Microélectronique et de Nanotechnologie, Université de Lille, Villeneuve d'Ascq, France
[2] NECOTIS Research Group, Université de Sherbrooke, Sherbrooke, QC, Canada
[3] Lille Neuroscience & Cognition, Université de Lille, Lille, France
[4] Laboratoire Nanotechnologies Nanosystemes (LN2)-IRL3463, CNRS, Université de Sherbrooke, INSA Lyon, Université Grenoble Alpes, Ecole Centrale de Lyon, Sherbrooke, J1K0A5 Quebec, Canada

Corresponding authors:
Sean Wood, assistant professor: sean.wood@usherbrooke.ca
Alexis Mélot, PhD student: alexis.melot.etu@univ-lille.fr

## Abstract

Spike sorting is a crucial step in decoding multichannel extracellular neural signals, enabling the identification of individual neuronal activity. A key challenge in brain-machine interfaces (BMIs) is achieving real-time, low-power spike sorting at the edge while keeping high neural decoding performance. This study introduces the Neuromorphic Sparse Sorter (NSS), a compact two-layer spiking neural network optimized for efficient spike sorting. NSS leverages the Locally Competitive Algorithm (LCA) for sparse coding to extract relevant features from noisy events with reduced computational demands. NSS learns to sort detected spike waveforms in an online fashion and operates entirely unsupervised. To exploit multi-bit spike coding capabilities of neuromorphic platforms like Intel's Loihi 2, a custom neuron model was implemented, enabling flexible power-performance trade-offs via adjustable spike bit-widths. Evaluations on simulated and real-world tetrode signals with biological drift showed NSS outperformed established pipelines such as WaveClus3 and PCA+KMeans. With 2-bit graded spikes, NSS on Loihi 2 outperformed NSS implemented with leaky integrate-and-fire neuron and achieved an $F_1$-score of 77% (+10% improvement) while consuming 8.6 mW (+1.65 mW) when tested on a drifting recording, with a computational processing time of 0.25 ms (+60 µs) per inference.

Keywords: Spiking neural network, Sparse Coding, Spike Sorting, Unsupervised Learning, Neuromorphic Computing.

## 1. Introduction

Brain-machine interfaces (BMIs) are critical in bridging communication between neural populations and computational systems, and spike sorting algorithms play a central role in decoding neural activity [1]. By detecting and classifying action potentials (spikes) from recorded neural signals, these algorithms enable the identification of single-neuron (or single-unit) activity, making them indispensable for advancing BMIs and understanding neural circuits [2], [3]. The development of these algorithms follows recent innovations in electrophysiology, particularly high-density microelectrode arrays (HD-MEAs) that have an increased number and density of electrodes, allowing for more detailed neuronal activity recordings [4], [5]. With typical sampling rates between 10 and 30 kHz per electrode, such recording devices generate large volumes of data to process. Although some spike sorting software offers good performance even with high electrode numbers [6], [7], [8], and operates automatically [9], [10], they typically rely on offline processing using traditional power-intensive processors (CPUs, GPUs). These approaches require the transmission of raw signals to external devices, which is not feasible for embedded real-time BMIs. A solution is to perform on-chip, in situ spike sorting to reduce bandwidth demands by transmitting only sorted action potentials. This minimizes the power required to rapidly transmit high-dimensional neuronal data, thus reducing the risk of brain tissue damage due to heat dissipation [11], [12]. However, the challenge is to design a spike sorting solution that operates efficiently at the edge, balancing rapid data processing for real-time BMIs with the constraints of minimal power consumption and limited computational resources.

The spike sorting process typically consists of four stages: preprocessing, feature extraction/dimensionality reduction, clustering, and optionally, template matching. Preprocessing involves filtering, spike detection, windowing, and spike alignment to generate spatiotemporal spike waveforms (SW). Feature extraction then reduces the data dimensionality by retaining the most relevant features from these waveforms, then in the clustering step, extracted features are grouped into clusters representing individual neurons. Noisy recording channels complicate SW differentiation, a challenge for spike sorting pipelines. Techniques such as discrete wavelets [3], continuous wavelets [13] or independent component analysis (ICA) [14] have been used to enhance waveform separability. Clustering is then performed using unsupervised algorithms such as K-means [14], [15], superparamagnetic [3], Gaussian mixture models [16], density-based clustering like DBSCAN [7].

Recent progress in machine learning and deep learning has further advanced spike sorting. Artificial neural networks (ANNs), particularly convolutional networks and autoencoders, have shown strong performance on large-scale recordings [17]. Autoencoders, which learn compressed latent representations of waveforms, have proven effective for feature extraction [18], [19], [20]. In terms of energy consumption, these models can be deployed on low-power edge-AI processors (e.g. edge-Tensor Processor Units) [21] or simplified into shallow two-layer networks [22] to require less CPU computational resources, but they often require supervised training, introducing performance-energy trade-offs [21], [22].

To address the need for low-energy, unsupervised, real-time spike sorting, spiking neural networks (SNNs) offer a promising solution [23], [24]. These networks operate using sparse, spike-based communication and can be trained with bio-inspired Hebbian learning rules such as spike-timing-dependent plasticity (STDP) [25], or Oja's rule [26]. Notable examples include Werner et al.'s two-layer SNN with filter banks and lateral inhibition for real-time spike sorting of single-channel recording [27] and Bernert et al.'s approach using an attention mechanism to restrict learning to spiking events in tetrode recordings, but require numerous parameters to be trained per channel which complexify potential hardware implementation [28]. More recently, the NeuSort algorithm introduced an adaptive filter bank with Hebbian learning to handle signal non-stationarities such as drifting and new neurons. Despite improved adaptability, hardware implementation remains an open issue.

SNN-based spike sorting methods are compatible with neuromorphic hardware processors. These bio-inspired processors excel in parallel, high-speed computation while consuming significantly less power than traditional computing systems based on von Neumann architectures [29], [30]. Recent advances in neuromorphic computing, such as in-memory analog architectures using memristors [31], have led to SNN-based spike sorting solutions, but remain limited to processing single-channel recordings and have sensitivity to noise [27]. To our knowledge, no spike sorting solution has been implemented on digital neuromorphic chips such as Intel's Loihi 2 [32] and SpiNNaker [33].

In previous work [34], the Locally Competitive Algorithm (LCA) [35], a recurrent ANN with bio-inspired internal neuronal dynamics, was shown to outperform other classical feature extraction methods such as PCA [36], [37] and K-SVD [6], particularly in SW with low signal-to-noise ratios (SNR). The LCA network is a sparse code estimator, meaning that it learns to represent high-dimensional data as a linear combination of a small subset of basis vectors, promoting efficiency by ensuring that most coefficients remain zero. The sparse coding method has proven to be effective in filtering out noise, allowing key features to emerge more clearly which enhance further recognition tasks [38], [39]. Loihi 2 and SpiNNaker, support multi-bit spikes (up to 32-bits) to enhance SNNs performance [32]. Leveraging this flexibility, an LCA-based image recognition solution shown that increasing spike bit-width enhances accuracy [40]. This hardware capability provides a flexible trade-off between the performance of classical ANNs and the energy efficiency of SNNs, which is particularly relevant for designing BMIs across diverse scenarios.

This study introduces the Neuromorphic Sparse Sorter (NSS), an SNN designed for low-power, real-time unsupervised spike sorting on neuromorphic platforms. NSS leverages a spiking version of the LCA network to solve the extraction of features and clustering of multichannel SW. Optimized for tetrode recordings in this first study, the aim of NSS is to addresse key aforementioned challenges in spike sorting at the edge. In the following sections, we detail our methodology, experimental results in simulations and with runs on Loihi 2 neurocores and discuss the broader implications of this approach for advancing BMIs.

## 2. Materials and Methods

### *2.1 Real and simulated neural data*

The spike sorting performance of the proposed pipeline was measured on a total of 9 neural signals. Five of them are synthetic extracellular neural recordings generated using the *spikeinterface* Python library [8] largely used in electrophysiology to benchmark spike sorters. The others are real tetrode datasets (see Table 1 for dataset summary) .

- *Synthetic datasets:* First, NSS was tested and parametrized on the synthetic datasets. Five tetrode recordings were generated. The probe design was chosen to resemble, in terms of contact size and spacing, the silicon electrode arrays used to record the real tetrode dataset described below. The simulated recordings, sampled at 10 kHz, are populated with 5 neuron templates. The neuron templates were synthetically generated using the simulator's default model. The neurons were randomly positioned in a 3D space with a maximum depth of $35\mu m$ and an area delimited by the positions of the electrode pads with a margin of $5\mu m$. Five recordings were generated, to get a broader range of SNR, where the SNR of neuron $i$ is computed as $SNR_i = A_{max,i}/\sigma_b$ where $A_{max,i}$ is the maximum amplitude across all recording channels of the mean SW related to neuron $i$. $\sigma_b$ is the standard deviation of the background noise and was set to be in the range of $[8, 12]$ $\mu V$ of added Gaussian noise. This noise could be different for each channel to further match real experimental conditions. These synthetic tetrode datasets will be referred as $TS_{1\text{-}4}$ and $TS_0$ left aside for parameterization of proposed network. The number of detected spike timings within a 3 ms range of each other, or spike overlaps, represent 11.1%, 17.1%, 20.1%, 18.5%, 18.2% for $TS_0$ to $TS_4$ respectively (Table 1). Additional information related to the pairwise similarity of the bioneuron template shape is given in Figure S1.

- *Real-world datasets:* Afterwards, the performance of our pipeline was benchmarked on four real recordings from the hippocampus region CA1 of anesthetized rats, recorded by Buzsaki's Laboratory [40] and made publicly available through the Collaborative Research in Computational Neuroscience platform (crcns.org/data-sets/hc/hc-1). The recordings selected comprise a 4-minute-long tetrode extracellular signal along with the juxtacellular potential of one neuron. This latter signal gives the ground-truth spike timings of one neuron in the recording population to assess the performance. The signal is sampled at 10 kHz. From all the datasets available, those with a good signal were selected. Also, we ensured a broad range of SNR for the ground-truth neuron. To that extent, *d5331.01, d5611.04, d5611.05* and *d5611.06* were studied and will be referred to as $TR_{1-4}$ respectively. $TR_1$ is notable for exhibiting drift, meaning that the SWs gradually change shape, likely due to shifts in the alignment between the recording electrodes and the bioneurons.

## 2.2 Proposed Neuromorphic Spike Sorting Pipeline

### 2.2.1 Neural Signal Preprocessing

The raw multichannel neural recordings undergo three key preprocessing steps: filtering, spike detection, and spike alignment. First, the signals are band-pass filtered between 300 Hz and 3 kHz using a 3rd-order Butterworth filter to eliminate local field potentials, 50–60 Hz powerline noise, and other high-frequency electrical noise. Spike detection is then performed using a classical thresholding approach, where spikes are identified at five times the Median Absolute Deviation (MAD), a robust estimator of background noise in electrophysiological data [41]. The detection performance of this method in terms of precision and false positive rate (FP-rate) is better than the Nonlinear Energy Operator [42] used by previous low-power spike sorting approaches [12], [43], notably NeuSort [44] (Table 1 and Fig. S2 for more details). For each suprathreshold window indicating a spike event, only the timestamp of the largest peak across all recording channels is retained. This thresholding method is widely adopted in spike sorting due to its low computational demands [45]. Finally, a 3-millisecond window centered around the largest peak across channels is extracted. The resulting SW are then flattened as vectors of dimension 120 (4 channels of 30 samples each) to form the input data of NSS.

In the proposed pipeline, the term "spike" can refer to different phenomena, so precise terminology is used to avoid ambiguity. A SW refers to a processed putative spike detected from the extracellular multichannel recording. A plain spike represents the firing of an artificial neuron within the NSS network. Additionally, the terms bioneuron and unit specifically refers to the neurons being analyzed, real or simulated, as distinct from the artificial neuron models in the NSS network.

### 2.2.2 Sparse Coding with the LCA

Sparse coding is inspired by the behavior of neurons in the primary visual cortex (V1) [46]. This method, when applied to an input $\boldsymbol{x} \in \mathbb{R}^L$ generates a simpler representation in the form of a sparse vector $\boldsymbol{a} \in \mathbb{R}^M$, where most coefficients are zeros. The signal is approximated as a linear combination $\hat{\boldsymbol{x}} = \boldsymbol{D}\boldsymbol{a}$, where $\boldsymbol{D}$ is a dictionary composed of $M$ column vectors $d_m \in \mathbb{R}^L$ also called *atoms*. Typically, $M \geq L$, and the dictionary is then said to be overcomplete. The sparse representation $\boldsymbol{a}$ for a given dictionary is found by solving the Least Absolute Shrinkage and Selection Operator (LASSO) optimization problem (Eq1). This problem could also be referred as the Basis Pursuit Denoising problem [41]. The sparse representation $\boldsymbol{a}$ for a given dictionary is computed by balancing the trade-off between reconstruction accuracy and the sparsity of the solution.

$$\min_{\boldsymbol{a}} \left( \tfrac{1}{2} \|\boldsymbol{x} - \boldsymbol{D}\boldsymbol{a}\|_2^2 + \lambda \|\boldsymbol{a}\|_1 \right) \quad (1)$$

The problem is solved by minimizing the residual error between the input and the sparse reconstruction in the form of the Mean-Squared Error in (Eq1). The second term is a cost function on the sparse vector that ensures the vector is sparse, usually the $l_1$-norm is used so the problem is convex and a unique solution can be found [47]. The factor $\lambda$ is a trade-off parameter between these two terms and is data dependent. Various sparse code solvers have been proposed [48], [49], [50]. For this work, the LCA network was selected due to the possibility to convert it to a SNN using spiking neuron models [51], while still converging to a unique and optimal sparse code solution [47].

**Table 1:** Sum up of the synthetic and real world extracellular recordings used to assess NSS spike sorting performance.

| Type | Name | Characteristics | | Detection Metrics | | | |
|---|---|---|---|---|---|---|---|
| | | SNR | Spiking Rate (Hz) | Total Number of SW | Precision (%) | FP-rate (%) | Overlap rate (%) |
| Synthetic | $TS_0$ | 4.4, 8.3, 9.3, 12.0 | 6.7, 6.5, 6.5, 7.2 | 6511 | 95.0 | 19.4 | 16.5 |
| | $TS_1$ | 3.2, 5.5, 6.9, 12.4, 13.1 | 6.5, 6.1, 6.9, 8.4, 8.8 | 8036 | 86.7 | 15.9 | 17.1 |
| | $TS_2$ | 3.2, 4.1, 6.2, 10.5, 15.4 | 7.5, 8.4, 7.7, 8.4, 8.6 | 8303 | 80.8 | 20.8 | 20.1 |
| | $TS_3$ | 4.9, 5.4, 6.3, 10.1, 12.0 | 7.0, 8.3, 8.7, 9.1, 8.8 | 9883 | 94.8 | 11.5 | 18.5 |
| | $TS_4$ | 3.5, 9.5, 10.1, 10.2, 11.5 | 7.1, 8.6, 9.0, 7.8, 7.0 | 8847 | 90.0 | 12.1 | 18.2 |
| Real-world | $TR_1$* | 8.0 | 3.5 | 3516 | 99.7 | - | - |
| | $TR_2$† | 5.9 | 2.2 | 2198 | 99.1 | - | - |
| | $TR_2$ | 4.4 | 0.8 | 2188 | 99.5 | - | - |
| | $TR_3$ | 4.3 | 0.9 | 2538 | 94.6 | - | - |

*$TR_1$ present a drift (Fig.6). † Duration of all recording is 240s except TR2 which is equal to 200s.

Spiking versions of LCA have been implemented on Loihi [30] and TrueNorth [52] neuromorphic chips to efficiently compute sparse codes from natural images. In their foundational work [35], Rozell et al. introduced the LCA neuron model as leaky integrators (LI) that dynamically adjust their membrane potentials to minimize the objective function (1) w.r.t. $\boldsymbol{a}$. The membrane potential of the LI neurons is governed by an ODE:

$$\tau \frac{d\boldsymbol{u}}{dt} = \boldsymbol{D}^T\boldsymbol{x} - \boldsymbol{u} - (\boldsymbol{D}^T\boldsymbol{D} - \boldsymbol{I})\boldsymbol{a} \qquad (2)$$

A given input sample $\boldsymbol{x} \in \mathbb{R}^L$, or in our case a SW, is presented multiple times for the neurons to converge to a stable output activation, by competing to activate through recurrent lateral inhibition connections: $W_r = -(\boldsymbol{D}^T\boldsymbol{D} - \boldsymbol{I})$. The number of iterations needed for the network membrane and sparse coefficient to converge is data dependent and varies with the neuron leak time constant. In our case, it is set to 200 iterations during the learning phase and then reduced to 32 when the dictionary has converged. The terms of the right-hand side of (2) represent the projection of the input onto the dictionary or the input bias at each iteration, the leak of the membrane potential $\boldsymbol{u}$ and the lateral inhibition originating from neurons whose membrane potential has surpassed a threshold $\lambda$: $\boldsymbol{a} = T_\lambda(\boldsymbol{u})$ where $T_\lambda$ is an activation function. Several functions have been proposed for $T_\lambda$ [35], the most used is the soft-thresholding or softshrink function that corresponds to the $l_1$-norm as the sparsity cost function in (1) [41]. As demonstrated in [35], this cost function on the sparsity of the sparse coefficients noted $C(\boldsymbol{a})$ is determined by the chosen activation function with the relation: $T_\lambda(\boldsymbol{u}) = \boldsymbol{u} - \lambda \frac{d\,C(\boldsymbol{a})}{d\boldsymbol{a}}$.

The LCA denoising capability is closely linked to the threshold selection of the activation function in LI neurons. Higher thresholds result in fewer neuron activations, leading to sparser representations of the input. While this can leave a higher residual error, it significantly reduces energy consumption, an important consideration for hardware implementations. Conversely, lower thresholds lead to more neuron activation, capturing finer details of the input but also retaining more noise. Achieving an optimal balance between these extremes allows the input to be denoised while preserving key features [34], [53].

### *2.2.3 Learning Rule*

The training process follows a two-step approach, similar to many sparse-coding solvers: first, the sparse-coding inference is performed using a fixed dictionary, and once the sparse coefficients are determined, the dictionary is updated.

Initially, the dictionary $\boldsymbol{D}$ can be set using wavelet basis functions or randomly selected input samples [54], [55]. However, studies have demonstrated that dictionaries learned directly from the input signal result in more effective representations in terms of both quality and sparsity [56], [57]. Consequently, after random initialization, the dictionary is learned through an update rule derived from the gradient of (1)

$$\Delta\boldsymbol{D} = \eta(\boldsymbol{x} - \boldsymbol{D}\boldsymbol{a}) \otimes \boldsymbol{a} + \mathbf{E} \qquad (3)$$

where $\otimes$ represents the outer product between $\boldsymbol{a}$ the sparse code and the residual error, $\eta$ is the learning rate. This learning rule is a direct gradient descent. The forward pass uses the quantized activation function to compute the internal dynamics during the iterations of each input, then the decoded activation is used for the learning process (Eq.3). This rule bears similarity to Hebbian learning, particularly in its causal form, where learning only occurs when a neuron spikes, reinforcing the connection between active neurons and the input that triggered their firing. This mechanism is advantageous as it aligns well with event-driven SNNs. To ensure stability and robustness in the training process an anti-Hebbian term is added to (Eq.3) in the form of a random zero-mean Gaussian noise matrix **E** of variance equal to 0.03.

### 2.2.4 Neuromorphic Sparse Coding-based spike sorting

The proposed spike sorting pipeline is illustrated in Figure 1. NSS (Figure 1.c) is a two-layer network where each layer corresponds to a single LCA network with recurrent connections. This architectural choice was motivated by a previous study that demonstrated the benefit of stacking multiple layers of sparse-coding solvers to learn hierarchical representations where higher layers combine dictionary elements from the previous layer to create new more global and abstract ones [58].

The first LCA, denoted as $LCA_1$, takes as input the flattened and preprocessed SW and extract its features in the form of a sparse code. At each iteration, only a subset of neurons in $LCA_1$ are active given the imposed sparsity constraint. The output of $LCA_1$ are fed directly into the second LCA ($LCA_2$). In this sense, the entire NSS behaves similarly to a conventional two-layer neural network.

The second layer serves as a clustering method. Its role is to assign the input to a class. The method used is a argmax-based labelling, where the class is defined as the index of the most active artificial neuron in $LCA_2$ at the end of the presentation steps of an SW. To minimize the memory footprint, the "winning" neuron is determined on the last 10 time steps of the SW presentation, which does not affect the dynamics of NSS. The code was made publicly available along with the tetrode simulated datasets[1].

The proposed NSS network is designed to perform unsupervised online learning on small batches of SW. It is trained in a layer-by-layer fashion. A scheduled learning rate was used to ensure a continuous adaptation throughout the recording, with strong learning phase for the first 60 seconds of recordings and a slower learning phase afterwards (Table2).

### 2.2.5 Graded Spiking Neuron Model

The LCA network, originally composed of LI neurons [35], replicates the continuous dynamics of biological neurons, but lacks the binary spiking output characteristic of SNNs. Spiking versions of LCA using leaky integrate-and-fire (LIF) neurons have been proposed for neuromorphic hardware, demonstrating significant power reductions [30],[59]. Motivated by the second-generation Loihi chip, which supports multi-bit spikes, an LCA-based image and video recognition solutions were developed leveraging this hardware capability. This approach enhanced performance while maintaining the temporal sparsity of SNNs and the spatial sparsity of the LCA's sparse code [60]. It used a method to quantize the activation function of LCA derived from the Temporally Diffused Quantizer (TDQ) [61]. Originally introduced by Voelker et al. to create "hybrid SNNs", the TDQ is used to quantize neuron activation into discrete steps while propagating quantization errors over time. This mechanism enables networks to leverage the high precision of artificial neural networks (ANNs) during training while smoothly transitioning to spiking regimes for inference. Also, the gradient descent training method remains valid since TDQ has a derivative equal to 1, as demonstrated in [61]. Their study showed a significant performance increase for bit-width close to 4-bits. TDQ serves as a generalized N-bit stepwise quantizer, transforming a non-spiking neuron's continuous output into discrete steps. The key mechanism of TDQ, represented as a block diagram in Figure 2, involves quantizing the neuron's activation at each time step, while propagating the resulting quantization error forward in time to minimize its impact on the network's performance. A rectified version of the softshrink function was used for NSS to avoid negative activation and thus simplify Loihi 2 implementation. The TDQ algorithm applied the latter activation function can be reformulated as follows:

$$\begin{cases} \tilde{a}_s(t) = T_\lambda(u(t), s) = \left\lfloor \frac{T_\lambda(u(t)) + v(t-1)}{s} \right\rfloor . s \\ v(t) = a(t) - \tilde{a}_s(t) \end{cases} \quad (4)$$

The key parameter $s$ determines the quantization step or "spike height" of the neuron output. It is defined by the ratio of the output range of the activation function and the chosen graded spike bit-width $N$: $s = 1/(2^N - 1)$. In our case, NSS outputs are bounded because SW are normalized to the unit norm to ensure stability. Also, the dictionary atoms (NSS weights) are normalized with $L_2$-norm to ensure that Eq.2 stays true [35], so in the end $|T_\lambda(.)| \leq 1$ is verified. The quantization error, noted $v$ in Equation 4, is the difference between the continuous output $a(t)$ and its discrete representation $\tilde{a}_s(t)$. As the parameter $s$ decreases the neuron behaves more like its original non-spiking form, while larger values increase the temporal sparsity of spiking, allowing the neuron to spike less frequently but still preserve the same average output. Thus, the TDQ allows for a flexible trade-off between sparsity and accuracy, making it possible to interpolate between classical 32-bit activation functions in ANNs and the discrete binary outputs of spiking neurons in SNNs as displayed in Figure 2. The TDQ is equivalent to the spiking integrate-and-fire neuron model for $s \leq 1$ without a refractory period when applied to the ReLU activation [61]. The rectified softshrink function is a ReLU-like function with the only difference that the threshold $\lambda \in ]0,1]$, preventing neurons with low membrane voltages from firing, thus acting as the sparsity coefficient. In the rest of the article, if a specific value of $N$ is used to run NSS, then it will be noted NSS-$N$bit.

[1] *https://github.com/NECOTIS/NSS-Neuromorphic-Sparse-Sorter*

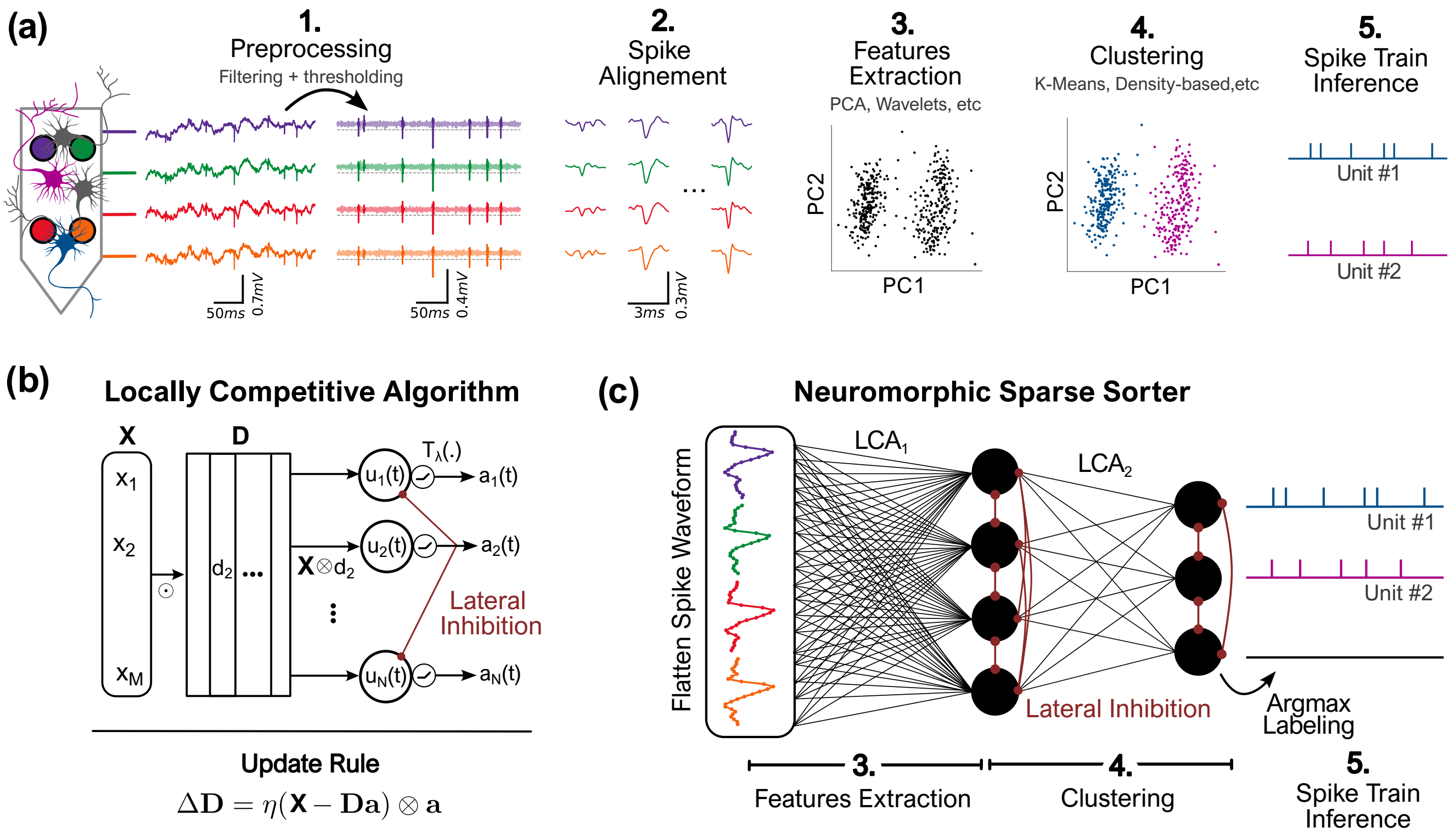

**Fig. 1: Overview of the Proposed Pipeline.** (a) Traditional spike sorting stages: 1. Filtering, whitening, and spike detection (threshold at 5× MAD); 2. Spike alignment by largest depolarization; 3. Feature extraction; 4. Clustering; 5. Spike train inference by combining spike timings and cluster labels. (b) LCA architecture: a sparse coding network with lateral inhibition. Each neuron's weight vector (or "atom") forms part of a dictionary, which is updated using a Hebbian-like rule after sparse inference convergence. (c) Proposed Neuromorphic Sparse Sorter: a two-layer LCA network for spike sorting. The first layer extracts sparse features, and the second layer classifies them.

## *2.3 Experimental Setups*

### *2.3.1 Hyper-parameter optimization*

The dictionary in the LCA network can either be fixed, using predefined basis functions or it can be learned directly from the input data. Studies have shown that learning the dictionary and optimizing LCA hyper-parameters from the input signal yields more effective representations in terms of both reconstruction quality and sparsity [34], [56], [57].

To ensure a consistent evaluation and avoid overfitting on the real-world dataset where the availability of ground-truth spike timing is limited, the hyper-parameters of NSS were optimized using an evolutionary genetic algorithm on a dedicated dataset $TS_0$ and then fixed for the other datasets (see Fig. S4 for a complementary sensitivity analysis). The optimization algorithm used was the Tree-structured Parzen Estimator selected by default by the *optuna* Python library, with its default search parameters.

Table 2 summarizes the hyperparameters of NSS used in this study. The dictionary size, beyond a certain point, has little impact on the sorting perfromance. In our previous work [34], we demonstrated that increasing the dictionary size to ten times overcomplete ($M = 10 \times L$) yields a performance gain of less than 0.1% in the $F_1$-score compared to $M = L$. Since a larger dictionary mainly increases processing time and energy consumption, a one-time overcomplete dictionary was chosen for $LCA_1$, while a dictionary with 10 atoms was used for $LCA_2$. The output layer (or dictionary) was chosen larger than the number of bioneurons to demonstrate that NSS performs well without prior knowledge of the recorded biological network. Figure S3 evaluates the impact of the dictionary size of $LCA_2$ on NSS performance and compares it with the $LCA_1$+KMeans pipeline. The comparison highlights a key advantage of $LCA_2$ over KMeans as a clustering method: $LCA_2$ does not require prior knowledge for parameterization. Once optimized, NSS is a rather small network with 130 neurons, and $L \cdot M_1 + M_1 \cdot M_2 + {M_1}^2 + {M_2}^2 = 30100$ synapses from which there are $L \cdot M_1 + M_1 \cdot M_2 = 15600$ learnable parameters, naming the forward weights. To estimate scalability, let's consider a hypothetical 64-channel recording with an input size $L = 1920$ and $M_1 = L$. In this case, the total number of synapses grows to ~7.6 million, which still fits within the capacity of a single Loihi 2 chip.

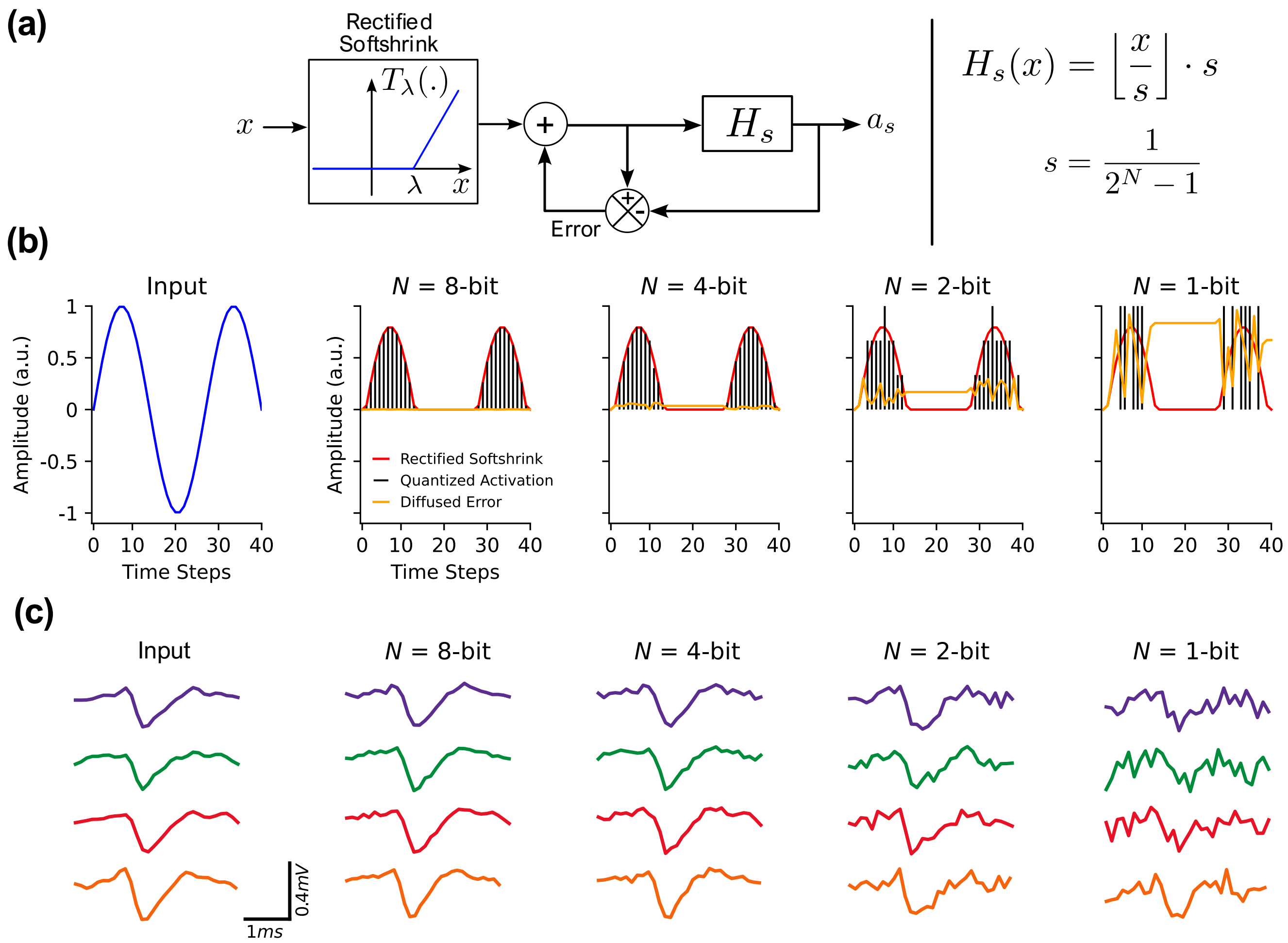

**Fig. 2: The TDQ algorithm applied on the LCA activation function.** (a) Block diagram of the TDQ algorithm applied to the rectified softshrink activation function used for LCA neuron. (b) Examples of the impact of varying N when $x$ from (a) is a sinus. (c) Sparse reconstruction from the output of the first LCA layer ($LCA_1$) of NSS from a detected SW when varying the bit-width of the graded spikes. $LCA_1$ is trained with $N$=1,2,4 or 8.

**Table 2:** NSS hyperparameters.

| Parameters | Description | Value |
|---|---|---|
| $M_1$, $M_2$ | Dictionary sizes/number of neurons per layer. | 120, 10 |
| $\boldsymbol{\lambda}, \boldsymbol{\lambda_{LIF}}$ | Firing threshold. | 0.03, 1.06 |
| $\boldsymbol{\tau}$ | Leak time constant. | 2ms |
| $\boldsymbol{\eta}$ | Scheduled learning rate. | 0.07 → 0.01 |
| $\boldsymbol{\Delta t}$ | Discrete time step. | 0.1ms |
| - | Scheduled number of time steps per SW. | 200 → 50 |

*2.3.2 Comparison with other sorters*

NSS was compared to two widely used spike sorting methods, PCA+KMeans (PCAKM) [62] and WaveClus3 [63], which serve as lightweight baseline methods in the literature. The methodologies were applied as follows:

- *PCA+KMeans*: The first three principal components (PCs) per channel were retained to reduce the dimensionality of SW. KMeans clustering was then applied, with $K = 5$ for simulated datasets and $K = 3$ for real datasets, in alignment with the number of sorted classes produced by NSS and prior studies using HC-1 datasets [18], [44]. SW from the first 60 seconds of recordings were used to train the PCA and initialize the KMeans centroids, while the remainder was used for evaluation.
- *WaveClus3:* A *Python* implementation of WaveClus3 was used with default parameters. The algorithm automatically optimized its temperature within the preset range.

Unlike NSS, both PCAKM and WaveClus3 operate in an offline fashion, requiring multiple iterations over the entire dataset of SW. In contrast, NSS processes data online, aligning with real-time spike sorting requirements. To further contextualize the performance of NSS within the landscape of existing spike sorters, we compared it against Tridesclous2, Spyking-Circus [7] and Kilosort [6], using their default parameter settings.

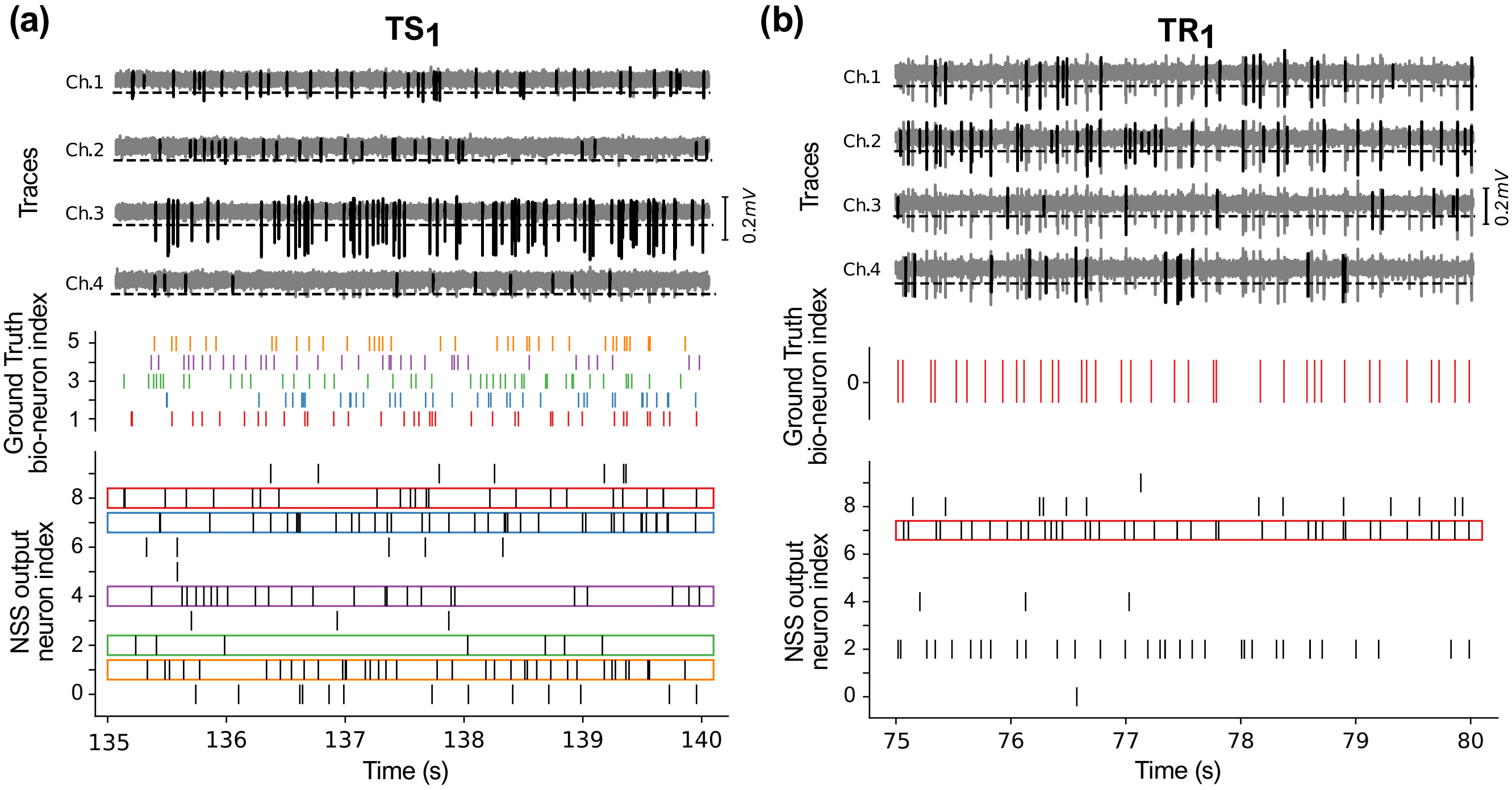

**Fig. 3: Examples of spike sorting using the NSS model on a simulated (TS) and a real tetrode (TR).** (a) Sorting result on $TS_1$ after NSS training has converged (after 60s of recording). Displayed for a 5-second snapshot of the recording. *Top*: Tetrode recording traces. *Middle:* Ground-truth raster of simulated bioneuron spiking events. *Bottom:* Inferred raster plot constructed by combining NSS sorted spike labels with detected spike timings from the thresholding phase. (b) Same as (a) for the real tetrode recording $TR_1$.

### *2.3.3 Evaluation Criteria*

The evaluation of the performance of our neuromorphic spike sorting algorithm starts by matching each ground truth neuron to NSS inferred units with the highest agreement score. The agreement score between each pair of ground truth and inferred units is calculated as the ratio of overlapping spikes in both rasters, considering a tolerance window of 1ms, to the total number of spikes in both spike trains. An illustration of the ground truth and inferred rasters is given in Figure 3. Each ground truth neuron is then matched to the inferred unit that achieves the highest agreement score.

To assess sorting accuracy, we use the $F_1$-score, a metric that balances precision and recall, offering a comprehensive measure of classification performance. Widely used in the spike sorting literature [28], [44], the $F_1$-score is calculated for each matched pair using the following formula:

$$F_1 = \frac{2 * TP}{(2 * TP + FN + FP)} \quad (5)$$

where TP, FN, and FP stand for true positives, false negatives, and false positives, respectively. For synthetic datasets, where the ground truth spike timings for all neurons are known, an $F_1$-score is computed for each neuron. In the real tetrode datasets, where only the spike timings of a single ground truth neuron are available, we calculate the $F_1$-score for this specific neuron only.

### *2.3.4 Neuromorphic Implementation*

The tests with a neuromorphic chip were performed using Intel Loihi 2 chips. These recent hardware versions allow graded spikes, which motivated the choice of the TDQ algorithm. First, NSS was trained offline on a CPU with $N$=8 on the first 60 seconds of $TS_1$ and $TR_1$. Then the frozen weights, time constant, and input SW detected after 60s for validation were converted to match the chip's fixed-point (32-bit integers) representation requirements. The energy and power consumption, were measured on the board *ncl-ext-og-05* and partition *Oheogulch*. The software version used to program the chip was *Lava 0.9.0*.

[2] *https://github.com/tridesclous/tridesclous*

The implementation of NSS on Loihi 2 required the programmation of a custom micro-programmed neuron model. The whole network used 2 neurocores out of the 128 available per chip. The I\O transfer time or off-chip communications took on average 117.5 ms to transfer a SW encoded into 32-bit graded spikes and read NSS outputs. This is a notable latency attributed to current hardware limitations expected to be corrected in future versions. All subsequent inference time measurements reflect only the neurocores computation time and exclude I\O latency.
No on-chip learning was performed in this study, as the current Loihi 2 architecture does not support layerwise learning rules required by NSS. The implementation of dictionary learning with LCA using local synaptic updates remains, to our knowledge, an open research problem [64]. Addressing this challenge falls outside the scope of the current work.

To benchmark performance and power consumption, we compared NSS with TDQ algorithm (NSS-TDQ) with a version using LIF neurons (NSS-LIF) on Loihi 2. The LIF threshold ($\lambda_{LIF} = 1.06$) was optimized experimentally on the $TS_0$ dataset using the aforementioned *Python* optimization library.

## 3. Results and Discussion

The performance of the NSS model was evaluated on both simulated and real tetrode recordings. Figure 3(a) and 3(b) illustrate examples of NSS spike sorting results over 5-second segments from a simulated ($TS_1$) and a real ($TR_1$) recording, respectively. NSS learns to sort tetrode neural signals in a fully online and unsupervised manner. Comparison of the inferred and ground-truth rasters demonstrates a strong alignment between matched spike trains (indicated by colored boxes), with minimal false negatives observed. On average, NSS requires processing approximately 2400 SW to reach stability for simulated datasets 1200 for real datasets. Given the simulated neurons' spiking rates (6–10 Hz), this equates to a convergence time of 62.5 seconds on average, with only a few hundreds of processed SW per bioneuron (Fig.4 (a)).

In the following sections, NSS sorting performances will be analyzed in depth across selected recordings, benchmarked against WaveClus3 and PCA+KMeans methods, and evaluated for hardware efficiency on the Loihi 2 neuromorphic chip.

### 3.1. Impact of quantization

The number of bits, $N$, chosen to encode the spike height in the TDQ algorithm (Figure 2(b)) significantly impacts the temporal sparsity of NSS. Temporal sparsity is defined as the average proportion of time steps during which NSS neurons remain inactive. For each SW presentation, this corresponds to 20 ms (200 time steps) during the strong learning phase and 3.2 ms (32 time steps) thereafter. When $N$ is low, neurons accumulate quantization error over multiple time steps before reaching a sufficient higher spike height, thus reducing their firing rate, and thus increasing temporal sparsity (Figure 2(b)) which theoretically would save energy. It is worth mentioning that this spike height mechanism differs from the all-or-nothing behavior of a standard LIF neuron, instead functioning as a multi-bit quantized approximation of NSS activation function. Higher values of $N$ not only reduce the quantization error but also improve detected SW distinguishability (Figure 2(c)), which enhances performance in the second layer of NSS. This layer, responsible for classification, receives more refined and denoised approximations of the input waveform, resulting in improved spike sorting accuracy. In Figure 4, the impact of varying $N$ on NSS spike sorting performance is analyzed. As expected, the overall number of spikes (regardless of height) emitted by NSS per input SW rises rapidly for $N$<8-bits, after which it plateaus (Fig.4 (c)). Additionally, the $F_1$-score gain relative to NSS-1bit results (noted NSS-1bit $F_{1\infty}$ in Fig.4) start stabilizing at $N$=2 or 3 for both simulated and real recordings.

**Table 3:** Power and time consumption of NSS on Loihi 2. Comparison of TDQ and LIF neurons.

| **Neural Recording** | **Neuron Model** LIF or TDQ Bit-Width | **Dynamic power** ($mW$) | **NSS Processing Time** ($ms$) | **Dynamic Energy** (μ$J$) | **Energy Delay Product** (μ$J.s$) | **$F_1$-score** (%) | |
|---|---|---|---|---|---|---|---|
| $TS_1$ | LIF | 5.20 | 0.19 | 1.01 | 191.9 | 75.3 | |
| | 1 | 6.89 | 0.28 | 1.93 | 540.2 | 74.7 | |
| | 2 | 6.76 | 0.33 | 2.26 | 745.8 | 80.2 | |
| | 4 | 12.26 | 0.34 | 4.07 | 1383.8 | 80.5 | |
| | 8 | 15.04 | 0.34 | 4.96 | 1686.4 | 82.1 | |
| $TR_1$ | LIF | 7.95 | 0.19 | 1.54 | 292.6 | 80.4* | 59.0† |
| | 1 | 7.11 | 0.25 | 1.76 | 440.0 | 83.7* | 66.7† |
| | 2 | 8.60 | 0.26 | 2.24 | 582.4 | 87.2* | 71.4† |
| | 4 | 12.79 | 0.27 | 3.30 | 891.0 | 87.5* | 74.1† |
| | 8 | 18.30 | 0.27 | 4.68 | 1263.6 | 87.5* | 74.1† |

* Computed on 100 SW before drift from NSS outputs on Loihi 2. † Computed on the last 100 SW.

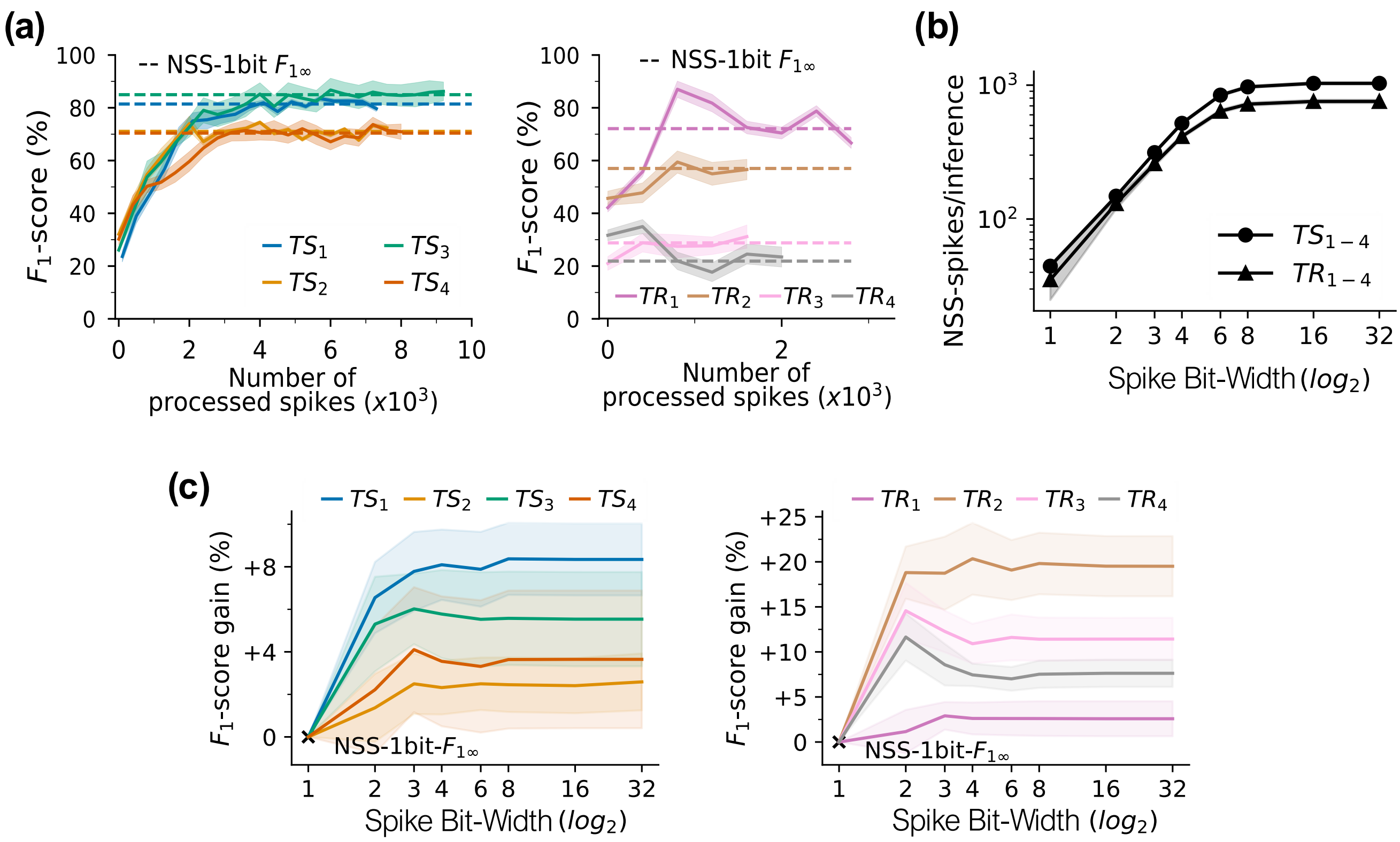

**Fig. 4: NSS sorting performance, efficiency and impact of graded spike bit-width.** The results are averaged over 20 trials with different random weight initializations; shaded areas show the 95% confidence interval. (a) NSS average sorting $F_1$-score evolution the simulated (*left*) and the real (*right*) datasets. The average $F_1$-score is computed by packets of 400 sorted spikes. (b) Average number of spikes emitted in NSS network to sort a SW. The spikes are counted regardless of their spike height. (c) $F_1$-score gain of NSS on the simulated (*left*) and the real (*right*) datasets relative to NSS-1bit performance after convergence noted $F_{1\infty}$ as it is the asymptote in (a). See Figure S6 for a detailed version of (c).

**Table 4:** Comparison of NSS with other spike sorting software.

| **Recording Name** | **Spike Sorter $F_1$-score (%)** | | | |
|---|---|---|---|---|
| | NSS-2bit | Kilosort4 | Trideclous | Spyking-Circus |
| $TS_1$ | 81.8 | 84.3 | 71.3 | 86.5 |
| $TS_2$ | 73.4 | 73.5 | 88.6 | 87.2 |
| $TS_3$ | 83.1 | 88.1 | 78.8 | 92.6 |
| $TS_4$ | 74.8 | 76.4 | 81.0 | 81.1 |
| $TR_1$ | 71.4 | N/A | 73.2* | 61.6* |
| $TR_2$ | 52.7 | 68.1 | 81.0* | 71.4* |
| $TR_3$ | 27.6 | N/A | N/A | N/A |
| $TR_4$ | 25.5 | N/A | N/A | N/A |

* $F_1$-score computed with precision and recall publicly available on Flatiron's Institute website SpikeForest. The initial release of Spyking-Circus was employed to process these two datasets, whereas subsequent analyses used the most recent version of the software.
N/A: Performance not available because of errors when running these sorters on these recordings.

This outcome suggests an optimal balance for choosing $N$ that maintains the number of spikes emitted by NSS to sort a single SW below 1000 (Fig.4 (b)) while benefiting from the performance gain of higher graded spike precision. Based on these findings, $N$=2 was selected for the remainder of the study as it captures this balance of energy savings, temporal sparsity, and classification performance gains. See supplementary Figure S5 for the detailed impact of $N$ on NSS sparsity.

### *3.2. Performance comparison*

The performance of our proposed sorting pipeline on simulated and real tetrode recordings is presented in Figure 5. In terms of $F_1$-score, NSS consistently outperforms both PCA+KMeans and WaveClus3 across most SNRs observed in these simulated datasets (Fig. 5 circles). NSS effectively handles varying noise levels and signal quality, benefiting from the demonstrated denoising ability of LCA [19], [41], [53]. On real datasets, NSS again shows competitive performance (Fig. 5 triangles). In terms of robustness to drift NSS outperforms PCA+KMeans but WaveClus3 is better, and we notice that on average NSS shows more robustness towards high overlaps rates.

The Figure 6 illustrates the drift of $TR_1$, with detected spikes from the ground-truth-labeled neuron changing shape over time. As a result all three methods exhibit a decline in $F_1$-score over time. Despite operating offline, WaveClus3 showed a decline of performance over time. However, NSS, which learns in an unsupervised online fashion unlike the other selected methods only slightly underperforms in $F_1$-score over the entire dataset compared to WaveClus3. Addressing drift correction is left for future work.

Apart from drift adaptation, long-term spike sorting solutions also face challenges due to heterogeneous firing rates, as some bioneurons may exhibit intermittent activity. This effect is observed in the $TR_1$ recording trace (Fig. 6(a)) where the ground truth bioneuron shows silent period during the strong learning phase of NSS (<60s) and after. This did not have a detrimental effect on the performance of NSS at all, it still stabilized after it has processed enough SW per bioneuron. The impact of newly firing bioneuron after network convergence has not been observed in real datasets or examined in simulations. This phenomenon is known to be a challenge for sparse code solvers because it requires an increase of dictionary size and thus more computations [65], this will be a focus of future work. Despite this limitation, NSS demonstrates competitive performance in controlled settings. However, its accuracy remains lower compared to more resource-intensive spike sorting methods such as Tridesclous, Spyking-Circus [7] and Kilosort [6] (Table 4). These algorithms benefit from more computationally complex pipelines, with steps dedicated to handle overlapping spikes [7] and drift [6].

### 3.3. NSS time/energy consumption on Loihi 2

Figure 7 shows how NSS is mapped onto two neurocores of the Loihi 2 chip, with each core hosting one network layer. The first and second layers use approximately 77 KB and 6 KB of SRAM, respectively, for neuron states and 12-bit synaptic weights. The efficiency of NSS on Loihi 2 was measured on datasets $TS_1$ and $TR_1$ and compared to a version of NSS implemented with LIF neurons. The default LIF neuron model proposed on Loihi 2 chip was used. Table 3 summarizes the comparative performance of these NSS versions on Loihi 2.

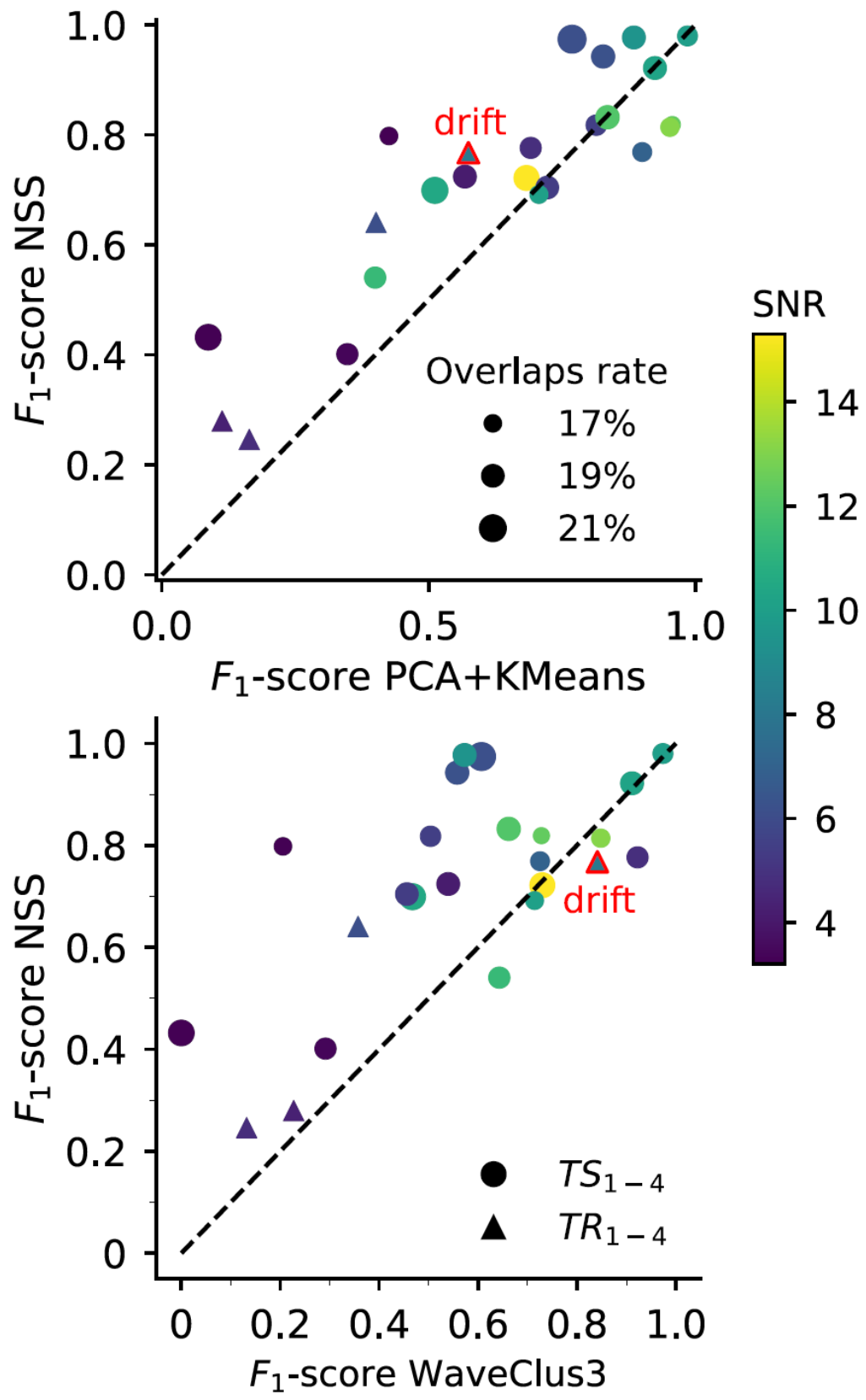

**Fig. 5: Benchmarking NSS sorting performance against WaveClus3 and PCA+KMeans on simulated and real tetrode datasets.** The results are averaged over 20 trials with different random weight initializations. The first 60s of the recordings are used for training the methods. The pairwise $F_1$-score comparison of the three sorting pipelines are plotted. Each colored dot represents the sorting $F_1$-score for a bioneuron, either simulated (circle) or real (triangle), with color indicating its associated SNR computed on the extracellular recording.

In the table, the power consumed by NSS is characterized by the dynamic power, whereas the static power measured on average at 450 mW is related to the hardware basis function that cannot be controlled. In preliminary CPU-based experiments, NSS-2bits demonstrated a favorable trade-off, reducing the total spike count and significantly boosting the sorting $F_1$-score (see Fig. 4). As anticipated, increasing spikes bit-width led to a higher dynamic power consumption for NSS-TDQ measured on Loihi 2, alongside notable improvements in the sorting $F_1$- score. For instance, for the $TR_1$ dataset, NSS-2bits, compared to NSS-LIF increased the energy-delay product (EDP) from 292.6 µJ·s to 582.4 µJ·s (or 1.99 mW/channel and 2.15 mW/channel respectively) while raising the $F_1$-score from 59.0% to 71.4% for the last 100 waveforms.

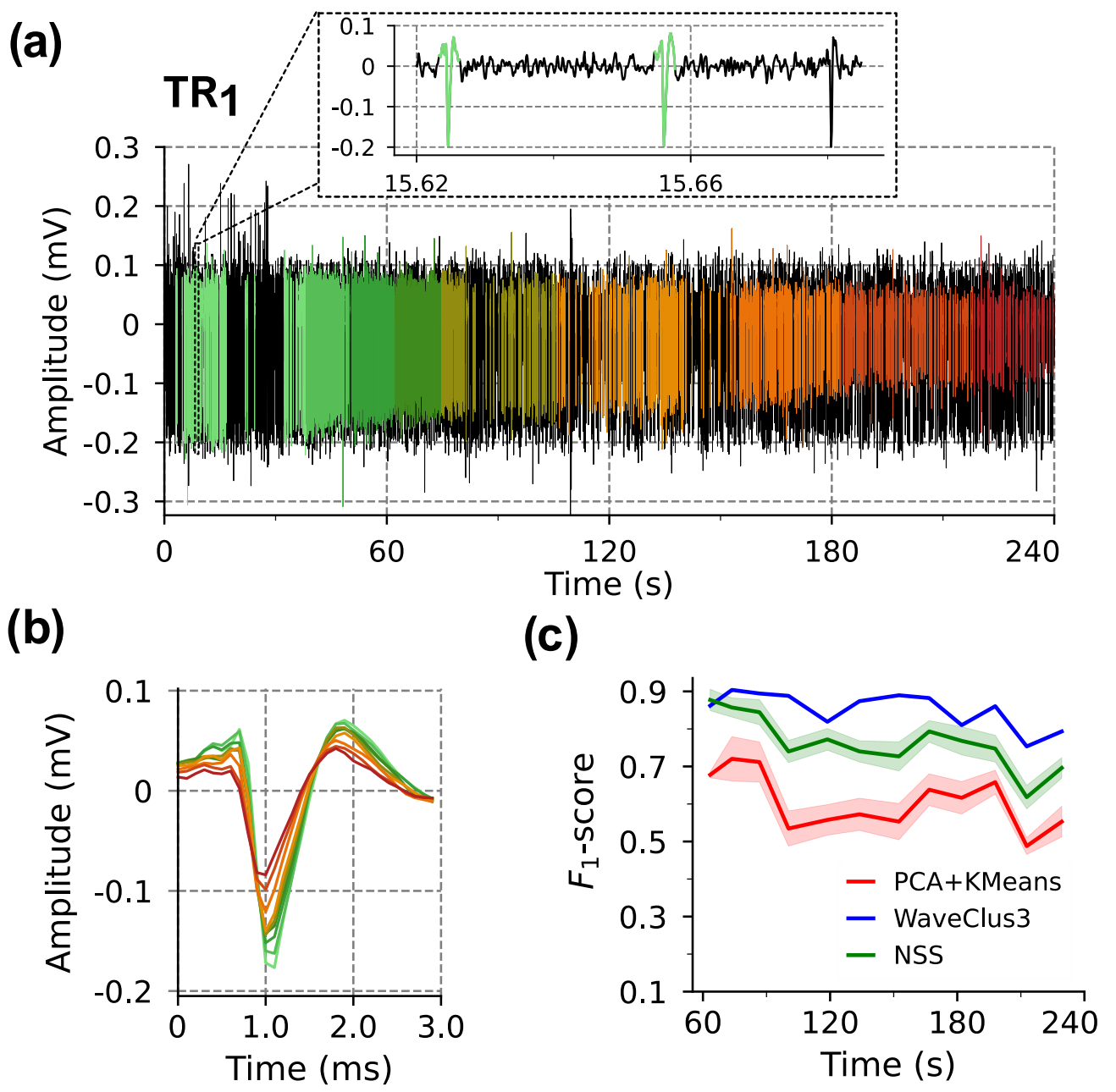

**Fig. 6: Robustness to drift on real dataset of NSS-2bit (CPU experiments).** (a) Extracellular recording trace for channel 1 of $TR_1$. There is a ground truth SW amplitude attenuation due to drift in $TR_1$. Spikes are highlighted by packets of 100 SW. (b) Average waveform over 100 SW. (c) $F_1$-score comparison of the three methods on $TR_1$. Average $F_1$-score and 95% confidence interval calculated at every 200 processed spikes. Since the first 60 seconds are used for training PCA and initializing KMeans clusters, the results are plotted starting from that time.

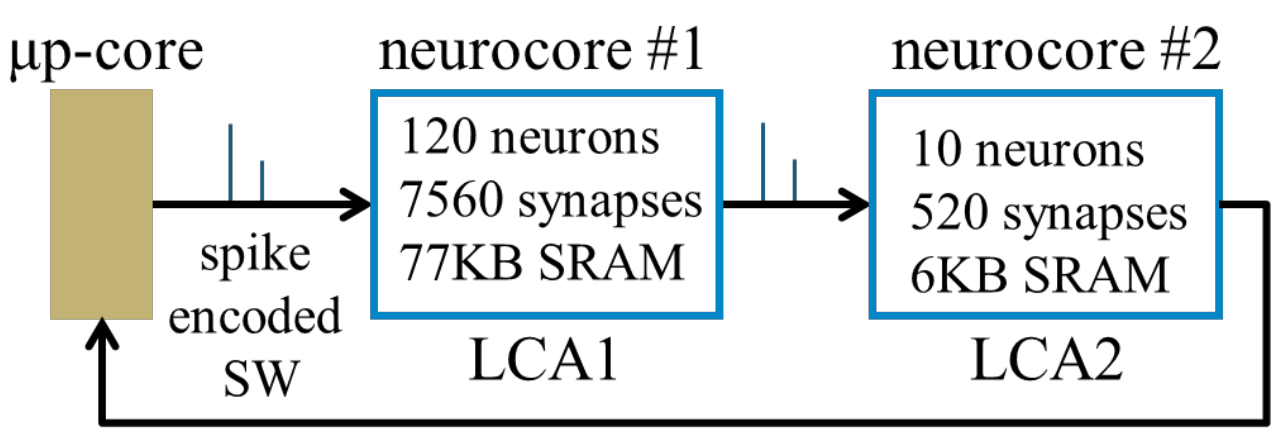

**Fig. 7: Hardware mapping of NSS on Loihi 2.** Diagram showing NSS implementation across two neurocores on a Loihi 2 chip, illustrating the system's hardware organization.

The measured dynamic powers to sort 4-channel SW remain as expected above the power consumed by spike sorting ASICs processors that go as low as 0.175uW/ch [12]. A recent study compared some spike sorting processors [18], which process multiple single-channel waveforms in parallel to optimize the workload. In our case, NSS uses 40% of the two allocated neurocores, which suggests further optimization could be done and that in the same way as the previously cited study, more recording channels could be processed in parallel. Otherwise, the inference time per SW remained below 0.4 ms, demonstrating the suitability of NSS for real-time processing of tetrode recordings. This processing time does not include the input/output communication latency to the neurocores and is measured by loading a flattened SW directly onto the neurocores at the start as an input bias to the neuron model.

These findings underscore NSS's potential, but further optimization is feasible. Currently, the online training is conducted off-chip with $N$=8 on a small batch of 16 SW. More energy savings could be achieved by performing on-chip learning with low output spike bit-width. Future work will focus on continuous adaptation to address drift in real time across diverse scenarios (e.g. fast, non-homogeneous, irreversible) and in high-channel-count recordings, aiming to enhance both efficiency and robustness.

Overall, using graded spikes and increasing the spike bit-width of NSS to $N$=2 showed good power-performance trade-offs for the tested recordings, but in the end the solution is conditioned by the available power at the edge. To address this, our proposed solution harnesses the flexibility in spike precision that the TDQ algorithm offers. NSS is a flexible sorting method adaptable to various BMI applications, where lower precision may be sufficient. Thus, NSS presents a versatile solution for unsupervised real-time spike sorting.

## Conclusion

In this study, we introduced the Neuromorphic Sparse Sorter (NSS), a compact two-layer neural network designed for online unsupervised spike sorting on neuromorphic hardware with minimal computational cost. NSS demonstrated superior performance compared to PCA+KMeans and WaveClus3 across a wide range of SNRs in both simulated and real recordings. NSS was implemented Intel's neuromorphic chip Loihi 2 that enables graded spikes up to 32 bits of precision. With a custom-made neuron model incorporating the TDQ algorithm, NSS demonstrated flexibility on the power-performance trade-off, by changing only one parameter, namely the spike bit-width, $N$. The use of graded spikes with height encoded using 2 bits instead of LIF neuron, raised the spike sorting $F_1$-score from 59.0% to 71.4% on a real recording after 4 minutes of slow drift. This increase of performance came at a marginal added dynamic power consumed from 7.95 mW to 8.60 mW. These findings point that NSS is an effective, low-power solution for spike sorting on neuromorphic platforms that propose graded spikes. The simplicity of changing the spike height to meet diverse real-time neural processing needs, make it a promising candidate for scalable deployment in brain-machine interfaces and similar applications.

## Acknowledgements

The project was funded by the MEIE research chair in neuromorphic computing and the ERC-IONOS project number 773228. The authors would like to express their gratitude to the Digital Research Alliance of Canada for providing access to computing clusters, and to the Intel

Neuromorphic Research Community for granting access to Loihi 2 chips and for their valuable support.

**Supplementary Materials**

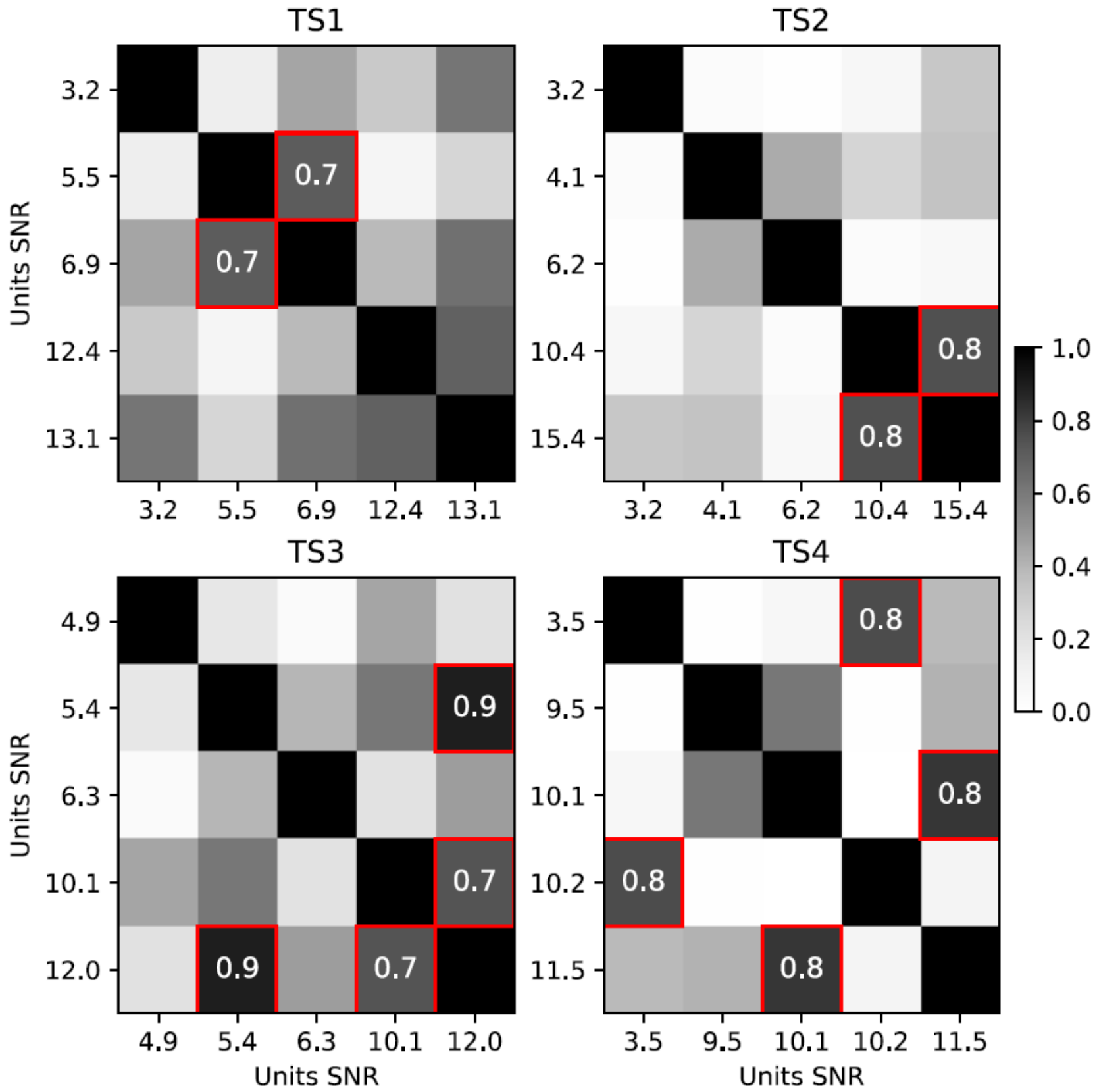

**Fig. S1: Pairwise cosine similarity matrices of bioneuron templates for the synthetic recordings.** Each matrix shows the cosine similarity between all pairs of ground-truth unit templates within a dataset. High similarity values (closer to 1) indicate overlapping or highly confusable spike shapes. The cosine similarity (CS) between two vectors A and B is computed as: $CS(A, B) = \frac{A \cdot B}{\|A\|\|B\|}$, with $\|A\|$ the l2-norm of the vector A.

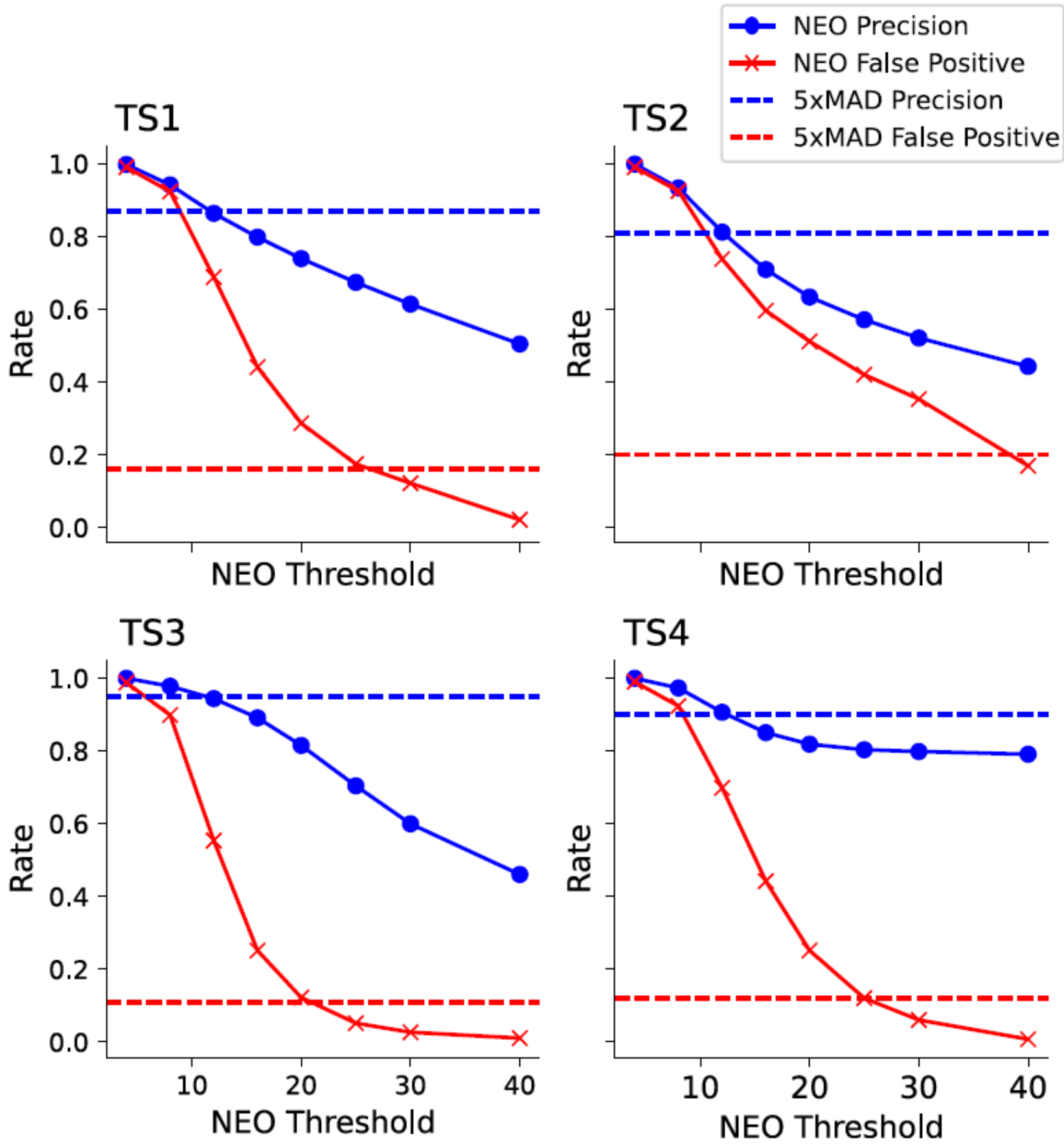

**Fig. S2: Comparison of detection method : 5xMAD vs. Nonlinear Energy Operator (NEO).** Performance comparison between the 5×MAD thresholding approach and the NEO method for signal detection. The figure illustrates differences in precision and false positive rate.

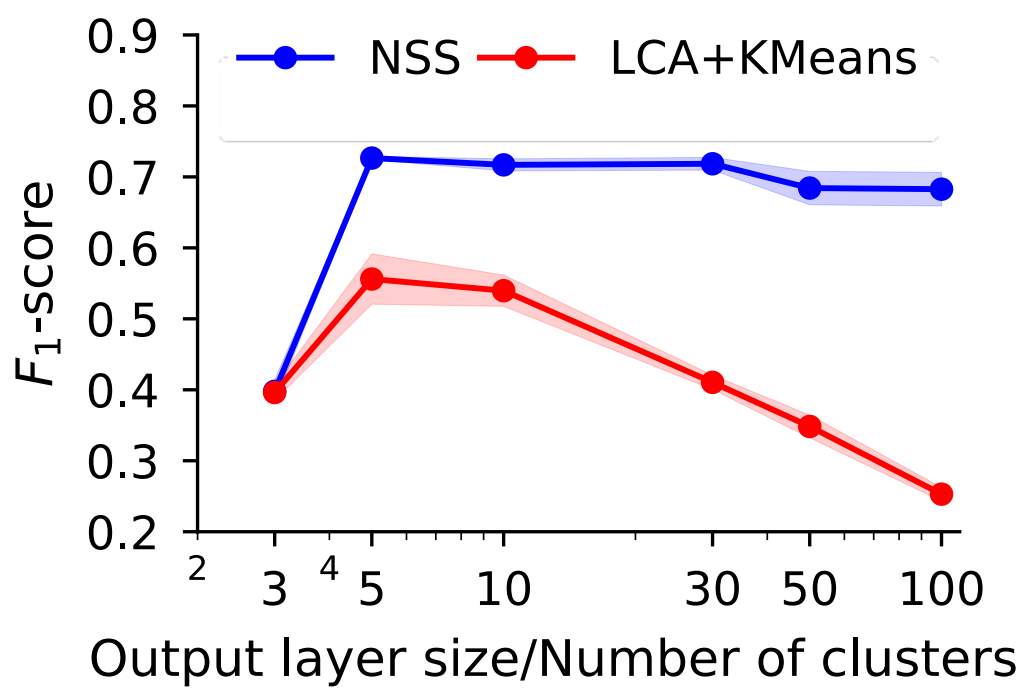

**Fig. S3: Comparison of NSS and LCA+KMeans - Impact of Output Layer Size on Performance.** This figure compares the performance of NSS and the LCA+KMeans pipeline on the TS1 dataset, focusing on the impact of varying the output layer size and the number of clusters in K-Means. The F1-score is averaged over the low-learning-rate phase of TS1. For K-Means, the first portion of the recording is used to fit the PCA, initialize its clusters, and train NSS. In the LCA+KMeans pipeline, the first layer of NSS (LCA1) is used for feature extraction, with K-Means replacing LCA2 to cluster the sparse codes generated by LCA1.The results demonstrate that NSS is robust to variations in the output layer size, in contrast to K-Means, which requires prior knowledge of the number of clusters to be defined. This robustness underscores that NSS does not require pre-defined parameters for clustering, making it a fully unsupervised spike sorting method.

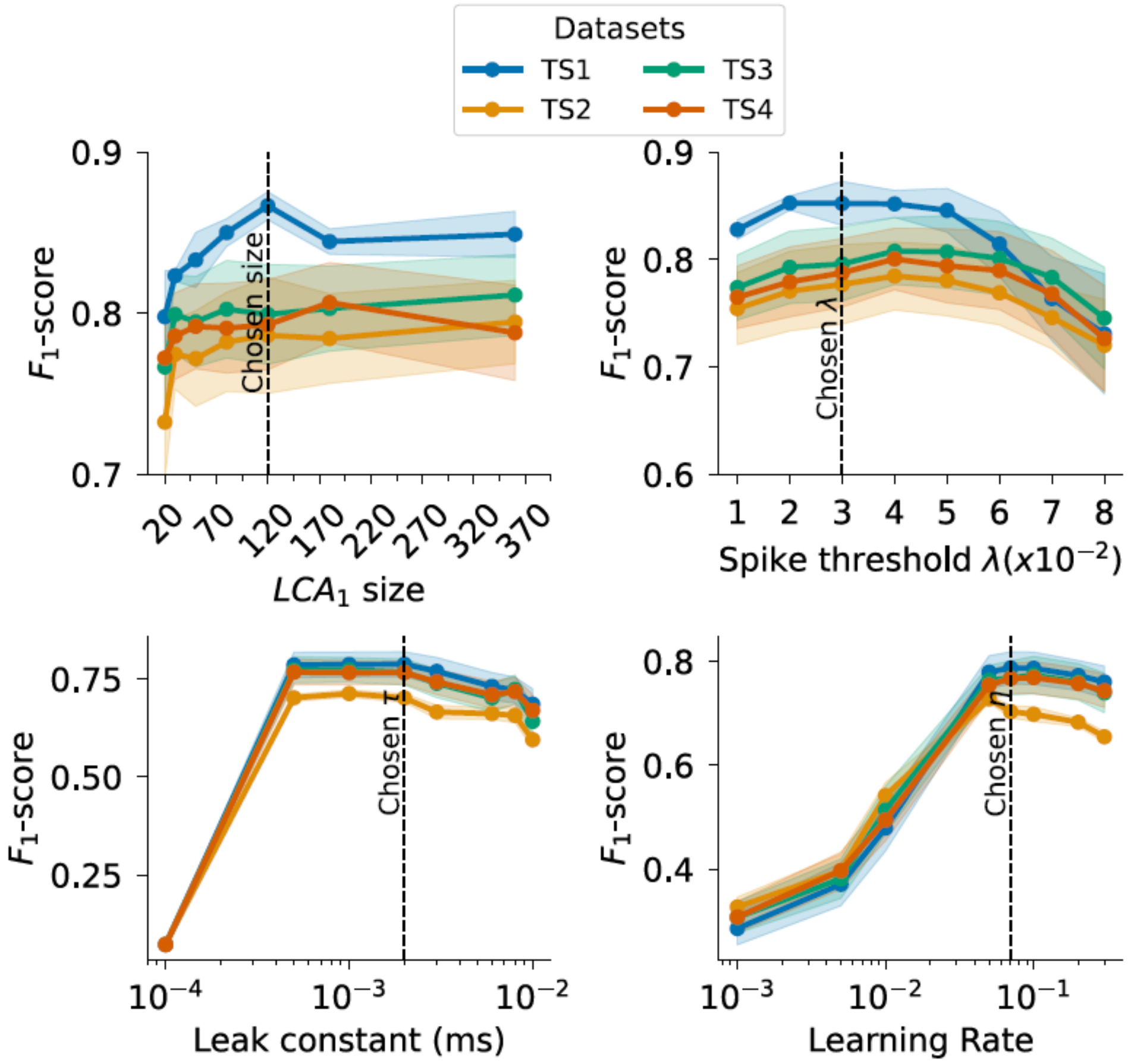

**Fig. S4: Sensitivity analysis of NSS performance to LCA hyper-parameters: LCA1 size, λ, τ, η.** F1-score of NSS evaluated on the synthetic dataset (TS1-4) as a function of LCA1 layer size, neuron activation threshold λ, leak time constant τ, and learning rate η. For each setting, 20 independent runs were performed to compute average performance and variability. Shaded areas represent ±95% confidence intervals.

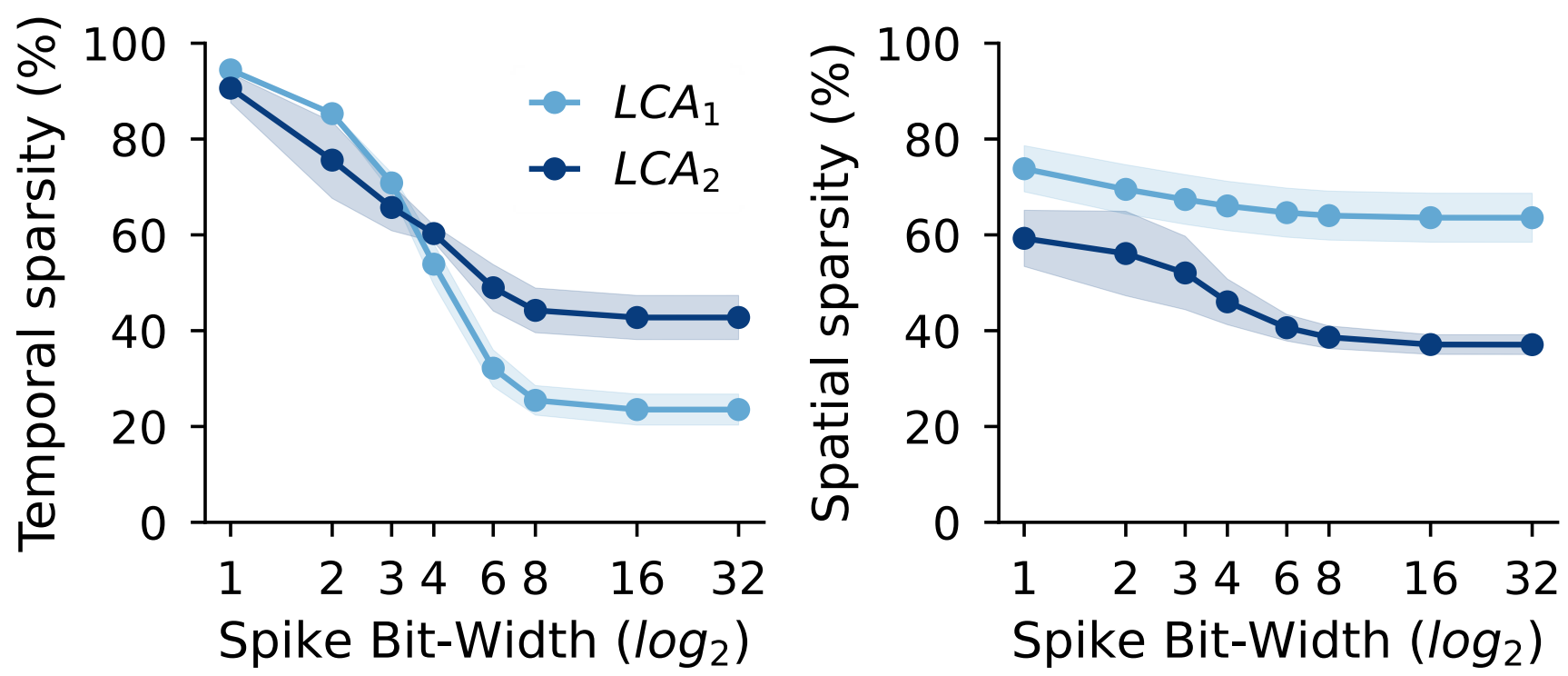

**Fig. S5: Effect of Graded Spike Bit-Width on Temporal and Spatial Sparsity in NSS**. This figure illustrates the impact of the graded spike bit-width on the temporal and spatial sparsity of the NSS network. Results are computed and averaged over the low-learning-rate phase of the TS1 dataset. Temporal sparsity is defined as the proportion of inactive time steps during the presentation of a spike waveform (SW), averaged across all neurons in the layer and multiple SWs. Spatial sparsity, measured using the L0L_0L0-norm, represents the average number of active neurons during SW presentations.The results show that temporal sparsity increases significantly as the spike bit-width decreases, indicating that NSS transitions toward behavior resembling spiking neural networks (SNNs). Conversely, spatial sparsity is only slightly affected; as the bit-width decreases, spatial sparsity exhibits a slight upward

trend. These findings highlight how reducing bit-width primarily promotes temporal sparsity, enabling a shift toward the energy-efficient characteristics of SNNs with minimal impact on spatial activity.

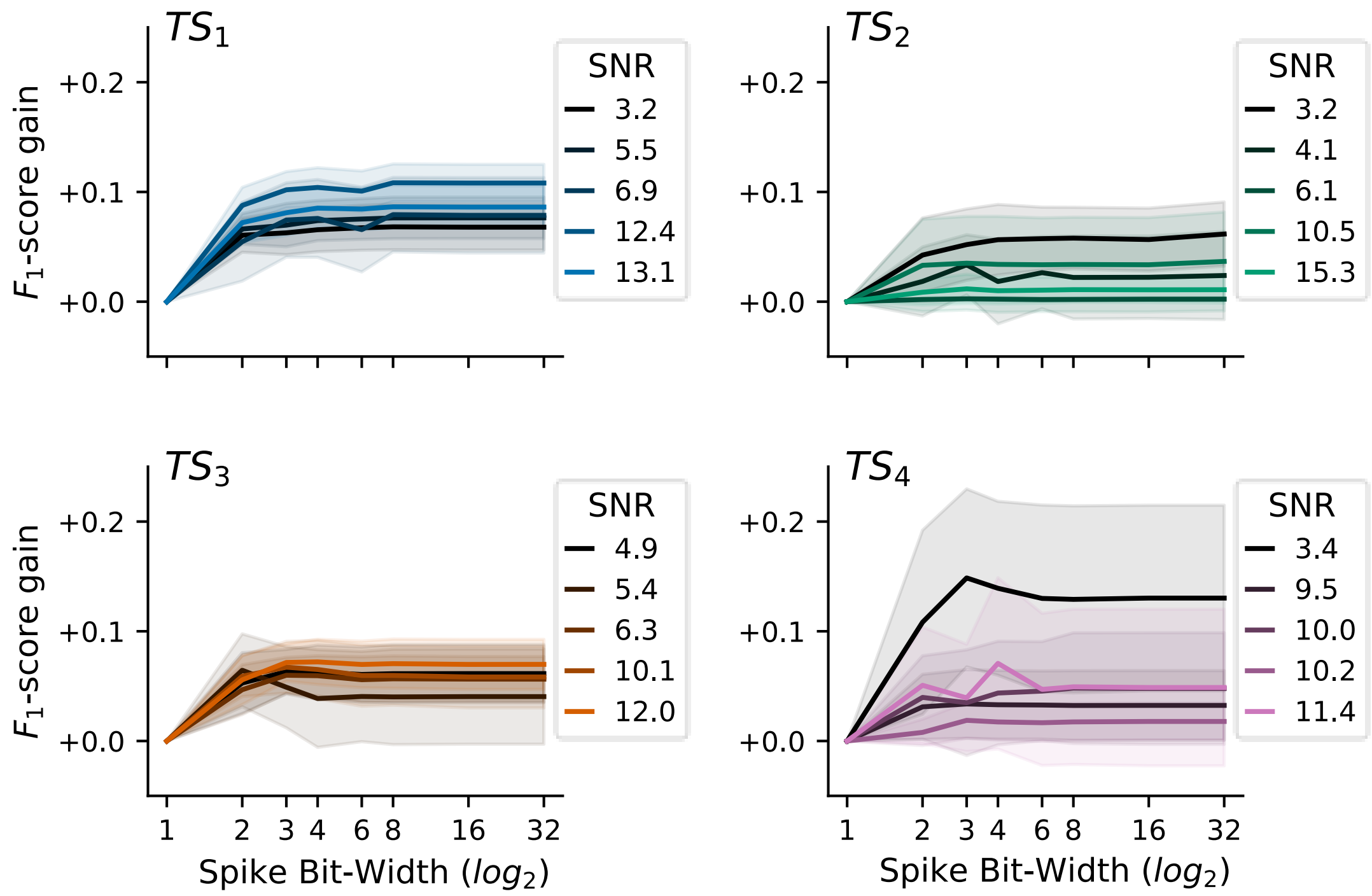

**Fig. S6: Effect of Spike Bit-Width on NSS F1-Score.** This figure provides a detailed analysis of the results summarized in Figure 4, focusing on the impact of varying spike bit-widths (using the TDQ algorithm) on NSS performance, as measured by the $F_1$-score for each bio-neuron in simulated datasets. Each panel corresponds to a specific dataset, consisting of five bioneurons identified by their signal-to-noise ratio (SNR) in the extracellular recording. The $F_1$-score improvement for NSS relative to its 1-bit graded spike performance after convergence, $F_{1\infty}$. , is shown.The results suggest that increasing spike precision has a greater effect on bioneurons with low SNR (e.g., in $TS_2$ and $TS_4$) or bioneurons with highly similar extracellular spike templates. For example, in $TS_1$, bioneurons with SNRs of 6.9 and 13.1 exhibit a pairwise cosine similarity of 0.75 between their spike templates, while in $TS_3$, bioneurons with SNRs of 12.0 and 5.4 show a similarity of 0.90. These findings indicate that higher precision graded spikes improve the ability to discriminate between noisy spike waveforms (SWs) and those with similar templates associated with different bioneurons. However, further in-depth analysis is required to confirm these observations regarding the influence of SNR and template similarity.

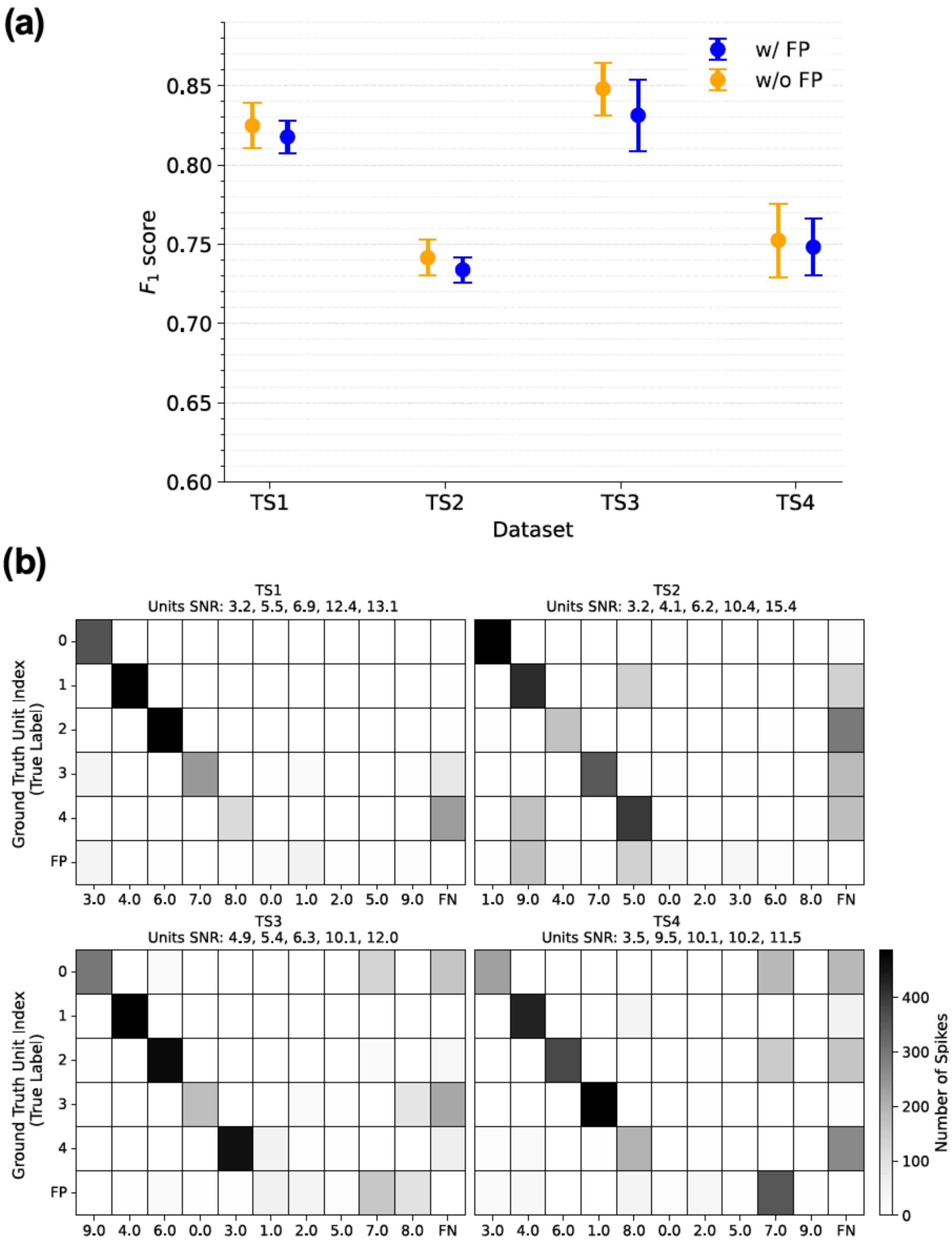

**Fig. S7: In-depth NSS performance evaluation.** (a) Visualization of the impact on NSS-2bit F1-score of SW falsely assigned as spiking event or FP cases. Average and 95% confidence interval over 20 runs for the case FP are manually removed from the simulated datasets. (b) Multi-labels confusion matrices of NSS-2bit for the synthetic datasets, including false positives and false negatives.